\documentclass{ecai}
\usepackage{graphicx}
\usepackage{latexsym}
\usepackage{mathtools}
\usepackage{amsmath}
\DeclareMathOperator*{\argmax}{argmax}
\usepackage[T1]{fontenc}
\usepackage[utf8]{inputenc}
\usepackage{multirow}
\usepackage{float}
\usepackage{enumitem}
\usepackage{makecell}
\usepackage{url}
\usepackage{hyperref}

\makeatletter
\newcommand{\specificthanks}[1]{\@fnsymbol{#1}}

\makeatother

\begin{document}

\begin{frontmatter}

\title{Investigating the Learning Behaviour of In-context Learning:  A Comparison with Supervised Learning}

\author[A, B]{\fnms{Xindi}~\snm{Wang}\orcid{0000-0002-6548-3547}\thanks{Work done during an internship at Microsoft STCA.}}
\author[C]
{\fnms{Yufei}~\snm{Wang}\thanks{\setcounter{author}{1}Work done during an internship at Microsoft STCA.}}
\author[D]{\fnms{Can}~\snm{Xu}} 
\author[D]{\fnms{Xiubo}~\snm{Geng}}
\author[D]{\fnms{Bowen}~\snm{Zhang}}
\author[D]{\fnms{Chongyang}~\snm{Tao}}
\author[B, E, F]{\fnms{Frank}~\snm{Rudzicz}\orcid{0000-0002-1139-3423}\thanks{~~Senior author.}}
\author[A, B]{\fnms{Robert E.}~\snm{Mercer}\orcid{0000-0002-0080-715X}\thanks{~~Senior author.}}
\author[D]{\fnms{Daxin}~\snm{Jiang}\thanks{Corresponding Author. Email: djiang@microsoft.com.}}

\address[A]{University of Western Ontario, Canada}
\address[B]{Vector Institute for Artificial Intelligence, Canada}
\address[C]{Macquarie University, Sydney, Australia}
\address[D]{Microsoft Corporation, Beijing, China}
\address[E]{University of Toronto, Canada}
\address[F]{Dalhousie University, Canada}


\begin{abstract}
Large language models (LLMs) have shown remarkable capacity for in-context learning (ICL), where learning a new task from just a few training examples is done without being explicitly pre-trained. However, despite the success of LLMs, there has been little understanding of how ICL learns the knowledge from the given prompts. In this paper, to make progress toward understanding the learning behaviour of ICL, we train the same LLMs with the same demonstration examples via ICL and supervised learning (SL), respectively, and investigate their performance under label perturbations (i.e., noisy labels and label imbalance) on a range of classification tasks. First, via extensive experiments, we find that gold labels have significant impacts on the downstream in-context performance, especially for large language models; however, imbalanced labels matter little to ICL across all model sizes. Second, when comparing with SL, we show empirically that ICL is less sensitive to label perturbations than SL, and ICL gradually attains comparable performance to SL as the model size increases. 
\end{abstract}

\end{frontmatter}

\section{Introduction}
Recent advances in large-scale pre-trained language models (LLMs), such as GPT-3 \cite{NEURIPS2020_1457c0d6}, have led to an interesting emergent learning paradigm called \textit{in-context learning} (ICL). In the ICL paradigm, given a prompt that includes a list of few-shot training input-output data and a test input at the end, LLMs directly make a prediction conditioning on the prompt without any updates to their model parameters. This is in contrast with current standards in fine-tuning \cite{devlin-etal-2019-bert, radford2019language}, where model parameters are updated according to the gradients of training losses. Compared to supervised fine-tuning (supervised learning), ICL allows users to directly manipulate LLMs  with only language-based prompts and more modest computational resource requirements.

However, despite the advantages of ICL, it is still unclear how ICL learns  knowledge from the given prompts without updating its model parameters. Preliminary research  \cite{akyurek2022learning,garg2022can}  compared ICL with simple machine learning models, such as logistic regression and shallow neural networks. In this paper, we take a further step and investigate learning behaviour differences between ICL and supervised learning (SL). Specifically, we train three LLMs with the same training data via \textit{in-context learning} and \textit{supervised learning} separately and analyze their generated outputs. While SL is a well-established approach that uses labelled data to train models to make accurate predictions, ICL takes a different approach by leveraging the context of the text to learn from unlabeled data in order to improve the accuracy of the predictions. By comparing the performance of ICL and SL, we gain insights into the effectiveness and weaknesses of each approach. In addition, previous work on ICL has hinged upon clean and balanced data; however, in practice, these conditions are incredibly difficult and expensive to meet. In situations with unclean or imbalanced data, ICL may provide a more cost-effective approach to learning from  limited available data, and comparing it with SL can help understand its potential advantages and limitations. Inspired by previous work which uses perturbed data to investigate properties of deep neural networks \cite{IVANOVS2021228}, we apply \textit{label perturbations} (i.e., incorrectly annotated labels and imbalanced distributed labels) to the above training data and observe the corresponding performance changes in both types of learning paradigms. This differs from previous ICL research  \cite{pmlr-v139-zhao21c} which only used balanced training data with high-quality annotations.

\begin{figure*}[t]
\begin{center}
\includegraphics[width=\textwidth]{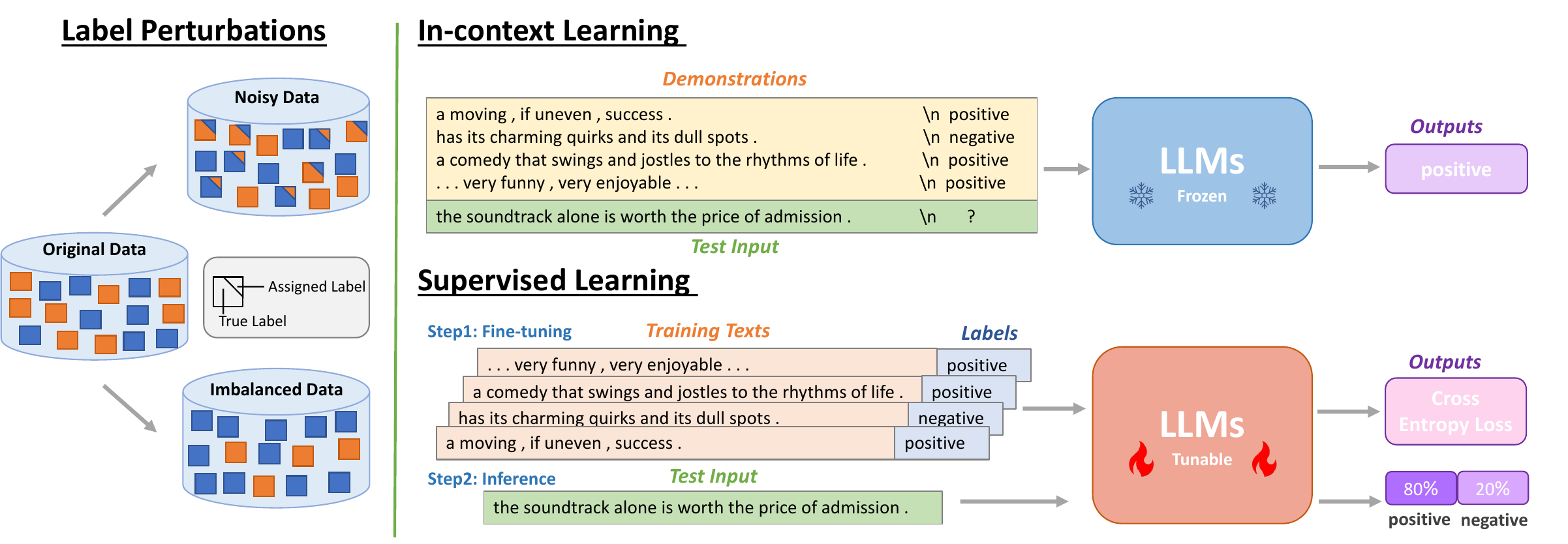}
\caption{A overview of our empirical studies. There are three main steps in our experiments. First, construct perturbed data from the original data. Second, we perform in-context learning: the demonstration examples consist of $k$ input-label pairs from the training data ($k = 4$ in this figure). Lastly, supervised learning: first fine-tune the LM using the same data as demonstrations in ICL and then use the fine-tuned LM for inference.}
\label{fig:1}
\end{center}
\end{figure*}

In this paper, we investigate experimentally the learning behaviour of in-context learning and compare it with supervised learning under label perturbations, as shown in Figure \ref{fig:1}. We compare model performance over six text classification datasets with different model sizes, and we evaluate the sensitivity of supervised learning and in-context learning on perturbed and clean data. 
We also provide a complementary study as to why the sensitivity of ICL differs with two label perturbation settings by calculating the attention scores on the labels. We provide the following major empirical findings on ICL under label perturbations: (1) gold labels are important to ICL and large language models are more sensitive to corrupted labels than smaller models; (2) label imbalance matters little to ICL across all model sizes. 
In comparing the two learning paradigms (i.e., ICL and SL), we observe the following: (1) in-context learning is less sensitive to label perturbations than supervised learning; (2) in-context learning gradually attains comparable performance to supervised learning as the model size increases.
The contributions of this paper are: 
\begin{enumerate}[nosep]
    \item We provide a new way of understanding the learning behaviour of ICL by using perturbed labels.
    \item We propose an explanation of the learning behaviour of ICL by using attention scores. 
    \item We are the first to perform a systematical comparison of the learning behaviours of ICL and SL under label perturbations.
\end{enumerate}

\section{Terminology}
\subsection{Problem Formulation}
In this paper, we explore in-context learning in downstream classification task $\mathcal{T}$ which includes a discrete output label set $\mathcal{Y} = \{y_{1}, y_{2}, ..., y_{L}\}$ and corresponding training data 
$\mathcal{D}_{\mathcal{T}} = \{(x_{i}, y_{j})\}, i = 1, 2, ...,N$, where $x_i$ is the input text, $y_j \in \mathcal{Y}$, and $N$ is the number of input-output pairs in $\mathcal{D}_{\mathcal{T}}$. Our goal is to 
learn a model $\mathcal{M}$ that predicts the correct $y_j$ using $x_i$ as input.

\paragraph{Supervised Learning}
In the traditional supervised learning setting, given a training instance $(x_i, y_j)$, we perform the parameter update for $\mathcal{M}$ based on the training loss, as follows:
\begin{align}
    \mathcal{L}(\mathcal{M}(x_i), \pmb{y}_{j}, \pmb{\theta}),
\end{align}
where $\pmb{y}_{j}$ is the one-hot representation for $y_j$ and $\pmb{\theta}$ are the parameters of $\mathcal{M}$. 

\paragraph{In-context Learning} 
Instead of training with gradient descent, \emph{In-Context Learning} directly estimates $y_i$ based on the input text $x_i$ and $k$ demonstration examples $\mathcal{C}={(x^c_1, y^c_1), \cdots, (x^c_k, y^c_k)}$. The final prediction $y_i$ is chosen from $\mathcal{Y}$, as follows:
\begin{equation}
    y_i = \argmax_j \mathcal{M}(y_j\, |\, \mathcal{C}, x_i),
\end{equation}
where $\mathcal{C}$ is represented by a concatenation of input-label pairs using pre-defined templates. 
We refer the reader to Table \ref{tab:8} for the detailed templates that are used for each dataset. 

\subsection{Learning with Label Perturbations}
Previous studies on \emph{in-context learning} mostly focus on the clean and high-quality data settings where the training data has high-quality annotations and roughly equal distributions over different labels in $\mathcal{Y}$. However, in many real-word scenarios, meeting such requirements could be challenging. Furthermore, as shown in \cite{IVANOVS2021228}, perturbed data could also be helpful in understanding the behavior of many neural networks. Thus, in this paper, we investigate how \emph{in-context learning} reacts to perturbed training data and compare such behavior with \emph{supervised learning}. Specifically, we consider two types of perturbations that simulate different levels of noise and degrees of class imbalance in real-world situations. 

\paragraph{Noisy Labels} In learning with noisy labels, we inject noise to form a label-corrupted training set $\tilde{\mathcal{D}}_{\mathcal{T}} = \{(x_{1}, \tilde{y}_{1}), (x_{2}, \tilde{y}_{2}), \dots, (x_{N}, \tilde{y}_{N})\}$, that is obtained from a noisy joint distribution over $\mathcal{X} \times \tilde{\mathcal{Y}}$, where $\tilde{\mathcal{D}}$ contains a combination of clean samples (whose original label is clean $y_{i} = \tilde{y}_{i}$) and noisy samples (whose original label is corrupted $y_{i} \neq \tilde{y}_{i}$). Our goal is to learn a competitive classifier by training on  corrupted data $\tilde{\mathcal{D}}_{\mathcal{T}}$. We consider uniform (or symmetric) noise with noise levels $\epsilon$.

The uniform noise flips a label from its true class to any other class with equal probability, $p_{\textit{uni}}$, that can be defined as:
\begin{equation} \label{eq:1}
    p_{\textit{uni}}(\tilde{y} = j | y = i) = 
     \begin{cases}
     1 - \epsilon, \textrm{for} \ i = j \\
     \frac{\epsilon}{L-1}, \textrm{for any} \ i \neq j 
     \end{cases}\hspace{-3.5mm},
\end{equation}
where $p_{\textit{uni}}(\tilde{y} = j \,| \, y = i)$ represents the 
probability of changing
the clean label $i$ to the noisy label $j$, and $\epsilon$ is the noise level. 
\begin{table*}[!t]
\resizebox{\textwidth}{!}{
\begin{tabular}{lllll}
\hline
\multicolumn{1}{c}{\multirow{2}{*}{\textbf{Dataset}}} & \multicolumn{1}{c}{\multirow{2}{*}{\textbf{Input}}} & \multicolumn{1}{c}{\multirow{2}{*}{\textbf{Label}}} & \multicolumn{2}{c}{\textbf{Format}} \\ \cline{4-5} 
\multicolumn{1}{c}{} & \multicolumn{1}{c}{} & \multicolumn{1}{c}{} & Model & Template \\ \hline
\multirow{3}{*}{MR} & \multirow{3}{*}{a moving , if uneven , success .} & \multirow{3}{*}{positive, negative}    & GPT2-Large &  Input\textvisiblespace Label\textvisiblespace\\
 & & & GPT2-XL & Input\textvisiblespace Label\textvisiblespace \\
&  &  & GPT-J & Input \textbackslash n Label \textbackslash n\textbackslash n\textbackslash n \\ \hline
\multirow{3}{*}{SST2} & \multirow{3}{*}{sentence: well worth revisiting as many times}                          & \multirow{3}{*}{ positive, negative} & GPT2-Large &  Input\textvisiblespace Label\textvisiblespace\\ 
 & & & GPT2-XL & Input\textvisiblespace Label\textvisiblespace \\
&  &  & GPT-J & Input \textbackslash n Label \textbackslash n\textbackslash n\textbackslash n \\ \hline
\multirow{6}{*}{RTE} &\multirow{6}{*}{\makecell[l]{Alleged terrorists today, killed Dolores Hinostroza, \\ the mayor of Mulqui district, shooting her five times. \\ implies: The mayor of Mulqui district was murdered \\ with a firearm. True or False?}} & \multirow{6}{*}{True, False} & \multirow{2}{*}{GPT2-Large} &  \multirow{2}{*}{Input\textvisiblespace Label\textvisiblespace}\\
& & & & \\
 & & & \multirow{2}{*}{GPT2-XL} & \multirow{2}{*}{Input\textvisiblespace Label\textvisiblespace} \\
& & & & \\
 & & & \multirow{2}{*}{GPT-J} & \multirow{2}{*}{Input \textbackslash n Label \textbackslash n\textbackslash n\textbackslash n} \\ 
& & & & \\ \hline
\multirow{6}{*}{CB} &\multirow{6}{*}{\makecell[l]{B: Right, you know, like In packaging A: \\ Yeah. B: and, uh, you know, just goodness. \\ A: Yeah, I don't think they do the packaging \\ at this plant, question: they do the packaging \\ at this plant. Entailment, contradiction, or neutral?}} & \multirow{6}{*}{entailment, contradiction, neutral} & \multirow{2}{*}{GPT2-Large} &  \multirow{2}{*}{Input\textvisiblespace Label\textvisiblespace}\\
& & & & \\
 & & & \multirow{2}{*}{GPT2-XL} & \multirow{2}{*}{Input\textvisiblespace Label\textvisiblespace} \\
& & & & \\
 & & & \multirow{2}{*}{GPT-J} & \multirow{2}{*}{Input \textbackslash n Label \textbackslash n\textbackslash n\textbackslash n} \\ 
& & & & \\ \hline
\multirow{6}{*}{AG-NEWS} & \multirow{6}{*}{\makecell[l]{Toshiba is taking back bad memory ZDNet's survey of \\ IT professionals in October kept upgrading hardware \\ at number two on the businesses radar throughout the year.}} & \multirow{6}{*}{Business, Science, Sports, World} & \multirow{2}{*}{GPT2-Large} &  \multirow{2}{*}{Input\textvisiblespace Label\textvisiblespace}\\
& & & & \\
 & & & \multirow{2}{*}{GPT2-XL} & \multirow{2}{*}{Input\textvisiblespace Label\textvisiblespace} \\
& & & & \\
 & & & \multirow{2}{*}{GPT-J} & \multirow{2}{*}{Input \textbackslash n Label \textbackslash n\textbackslash n\textbackslash n} \\ 
& & & & \\ \hline
\multirow{3}{*}{TREC} & \multirow{3}{*}{What actor came to dinner in Guess Who 's Coming to Dinner ?} & \multirow{3}{*}{\makecell[l]{Abbreviation, Description, Entity, \\ Person, Location, Number}} & GPT2-Large &  Input\textvisiblespace Label\textvisiblespace \\ 
 & & & GPT2-XL & Input\textvisiblespace Label\textvisiblespace \\
&  &  & GPT-J & Input \textbackslash n Label \textbackslash n\textbackslash n\textbackslash n \\ \hline
\end{tabular}}
\caption{A list of prompt templates we used for ICL. We show one example per task in the demonstration for illustration purposes. (Note: \textvisiblespace represents space, and \textbackslash n represents new line.) }
\label{tab:8}
\end{table*}

\paragraph{Label Imbalance} 
In practice, skewed data distributions arise in many datasets. In learning with label imbalance, the degree of class imbalance can be represented by the imbalance ratio. Given a balanced data distribution $\mathcal{D}$, let $r$ denote the ratio of class imbalance, $\mathcal{D}^{r}$ is the imbalanced data distribution for ratio $r$. That is, $r$ is the probability of the rarest class over that of the most frequent class:
\begin{equation}
    r = \frac{\min_{l \in \mathcal{Y}}\mathcal{D}(y = l)}{\max_{l\in\mathcal{Y}}\mathcal{D}(y = l)},
\end{equation}
where $0 \leq r \leq 1$. The dataset is balanced if $r=1$ and the dataset is heavily long-tailed if $r$ is small.
\subsection{Sensitivity to Label Perturbation}
Here, sensitivity is the degree to which the downstream classification performance changes when the model is subject to a fixed amount of label perturbation. Following Yoo {\em et al.}\ \cite{https://doi.org/10.48550/arxiv.2205.12685}, we measure the sensitivity using a linear regression analysis on the model performance against the perturbed rates:
\begin{equation}
    \begin{aligned}
       f_{\mathcal{M}} = \beta_{0} + \beta_{1}\mathcal{P}
    \end{aligned},
\end{equation}
where $f_{\mathcal{M}}$ is the performance of language model $\mathcal{M}$, and $\mathcal{P}$ is the perturbation rate. 
The scalar value $\beta_{1}$ is interpreted as the sensitivity measure. 

In general, the performance of a model drops with perturbation. The coefficient of the linear regression is usually negative, and a lower absolute coefficient value   indicates weak sensitivity. The intercept $\beta_{0}$ represents the performance of gold labels. 

\section{Experiments}
\subsection{Datasets}
We conduct experiments on six  datasets: binary sentiment classification with Movie Review (MR) \cite{pang-lee-2005-seeing} and SST-2 \cite{socher-etal-2013-recursive}; textual entailment using RTE \cite{10.1007/11736790_9} and CommitmentBank (CB) \cite{Marneffe2019TheCI}; topic classification using AGNews \cite{10.5555/2969239.2969312}; and question classification using TREC \cite{li-roth-2002-learning, hovy-etal-2001-toward}. Table \ref{tab:8} shows the default prompt format used for all datasets. For GPT2-Large and GPT2-XL, we separate the input and the label and each demonstration example with a space. For GPT-J, we separate the input and the label with a new line and each demonstration example with three new lines. The statistics of the datasets as well as the corruption rate and imbalance ratio for each dataset are summarized in Table \ref{tab:1}. 

We perturb the datasets in two ways: 
\begin{itemize}[nosep]
    \item Noisy labels: we corrupt labels by various amounts from 0\% to 100\%, with steps of 25\%, with uniform noise.
    \item Label imbalance: we construct label distributions with three levels of imbalance ratios: low (66\% to 100\%), medium (33\% to 66\%), and high (0\% to 33\%).
\end{itemize}

\begin{table}
\resizebox{\columnwidth}{!}{%
\begin{tabular}{c c c c c}
\hline
\multirow{2}{*}{\textbf{Dataset}} & \multirow{2}{*}{\textbf{Classes}} & \textbf{Number of} & \multirow{2}{*}{\textbf{Noise Level (\%)}} & \textbf{Imbalance}\\ 
 &  & \textbf{Testing} &  & \textbf{Ratio (\%)}\\
\hline
MR & 2 & 1066 & 100/ 75/ 50/ 25/ 0 & 100/ 45/ 7\\
SST-2 & 2 & 872 & 100/ 75/ 50/ 25/ 0 & 100/ 45/ 7\\
RTE & 2 & 277 & 100/ 75/ 50/ 25/ 0 & 100/ 45/ 7\\ 
\hline
CB & 3 & 56 & 100/ 75/ 50/ 25/ 0 & 83/ 43/ 10\\ 
AGNews & 4 & 7600 & 100/ 75/ 50/ 25/ 0 & 100/ 50/ 10\\
TREC & 6 & 1091 & 100/ 75/ 50/ 25/ 0 & 67/ 33/ 9\\ 
\hline
\end{tabular}
}
\caption{Details of the text classification datasets.}
\label{tab:1}
\end{table}

\subsection{Implementation Details}
We use 16 examples as demonstrations by default for both SL and ICL. We use 5 different random seeds and run all experiments 5 times. For SL, we first do grid search for hyper-parameter optimization on each dataset using 5 different seeds and then fine-tune the model with the optimum values of hyper-parameters per seed. 
We experiment with three language models, GPT2-Large \cite{radford2019language}, GPT2-XL \cite{radford2019language}, and GPT-J \cite{gpt-j}, ranging from 774 million to 6.7 billion parameters, all being decoder-only models. Evaluating on a comparable set of models with different sizes allows us to further investigate whether the observations depend on the model size. We report Macro-F1 scores for all tasks and compute the average score over seeds for each dataset. The code for reproducing our experiments is available at \urlstyle{same}\url{https://github.com/xdwang0726/ICL_LL}.
\section{Empirical Findings}
We intensively experiment to answer the following:
\begin{enumerate}[nosep]
    \item Does the ratio of correct labels matter to the performance of in-context learning?
    \item Is in-context learning less sensitive to noisy labels compared to supervised learning?
    \item Does the imbalance ratio affect in-context learning performance?
    \item Is in-context learning more sensitive to label imbalance compared to supervised learning?
\end{enumerate}
Tables \ref{tab:2} and \ref{tab:3} report the main results for in-context learning and supervised learning performance across three different models under noisy label and label imbalance settings, respectively\footnote{Detailed performance for each dataset can be found in Technical Appendix A at \urlstyle{same}\url{https://github.com/xdwang0726/ICL_LL/blob/main/ECAI_Supplementary_Document.pdf}. }. 
\begin{table*}[t]
\centering
\resizebox{\textwidth}{!}{
\begin{tabular}{cccccccccccc}
\hline
\multirow{3}{*}{\textbf{Dataset}} & \multirow{3}{*}{\textbf{LM}} & \multicolumn{10}{c}{\textbf{Noisy Level}} \\ \cline{3-12} 
 &  & \multicolumn{2}{c}{0} & \multicolumn{2}{c}{25} & \multicolumn{2}{c}{50} & \multicolumn{2}{c}{75} & \multicolumn{2}{c}{100} \\ \cline{3-12} 
 &  & SL & ICL & SL & ICL & SL & ICL & SL & ICL & SL & ICL \\ \hline
\multirow{3}{*}{Binary} & \multicolumn{1}{l}{GPT2-Large} & 51.6 \textsubscript{2.1} & 53.9 \textsubscript{16.9} & 45.4 \textsubscript{1.4}& 50.4 \textsubscript{11.2}& 46.9 \textsubscript{3.1}& 44.5 \textsubscript{3.7}& 45.8 \textsubscript{1.5}& 40.9 \textsubscript{2.9}& 44.1 \textsubscript{5.1} & 41.1 \textsubscript{1.8}\\
 & GPT2-XL & 49.4 \textsubscript{2.5} & 46.3 \textsubscript{8.7}& 45.0 \textsubscript{2.2}& 47.1 \textsubscript{6.4}& 46.6 \textsubscript{1.0}& 43.3 \textsubscript{6.5}& 45.7 \textsubscript{1.6}& 40.9 \textsubscript{5.8}& 43.0 \textsubscript{4.9}& 43.4 \textsubscript{6.2}\\
 & GPT-J & 47.2 \textsubscript{3.7} & 54.0 \textsubscript{13.4}& 43.8 \textsubscript{1.2}& 48.0 \textsubscript{13.7}& 45.5 \textsubscript{3.3}& 43.5 \textsubscript{3.6}& 41.9 \textsubscript{1.7}& 38.6 \textsubscript{2.3}& 38.3 \textsubscript{3.9}& 37.8 \textsubscript{1.7}\\ \hline
\multirow{3}{*}{Multi-class} & GPT2-Large & 42.6 \textsubscript{14.7}& 21.9 \textsubscript{12.1}& 33.6 \textsubscript{7.4}& 22.3 \textsubscript{13.6} & 25.0 \textsubscript{3.4}& 19.5 \textsubscript{12.4}& 18.5 \textsubscript{4.6}& 18.8 \textsubscript{13.1}& 13.8 \textsubscript{3.7}& 17.7 \textsubscript{11.0}\\
 & GPT2-XL & 41.9 \textsubscript{14.0}& 33.7 \textsubscript{9.7}& 33.2 \textsubscript{8.1}& 34.1 \textsubscript{8.0}& 24.5 \textsubscript{2.9}& 29.0 \textsubscript{4.1}& 20.3 \textsubscript{3.2}& 26.1 \textsubscript{4.3}& 13.0 \textsubscript{2.8}& 23.6 \textsubscript{4.9}\\
 & GPT-J & 42.4 \textsubscript{16.2}& 40.6 \textsubscript{11.4}& 38.5 \textsubscript{14.0}& 36.0 \textsubscript{10.2}& 28.1 \textsubscript{6.9}& 29.2 \textsubscript{3.5}& 19.1 \textsubscript{5.0}& 23.3 \textsubscript{6.8}& 12.4 \textsubscript{4.0}& 20.9 \textsubscript{7.9}\\ \hline
 \hline
 \multirow{3}{*}{AVERAGE} & GPT2-Large & 47.1 \textsubscript{10.6}& 37.9 \textsubscript{21.8}& 39.5 \textsubscript{8.0}& 36.4 \textsubscript{18.9}& 35.9 \textsubscript{12.4}& 32.0 \textsubscript{15.9}& 32.2 \textsubscript{15.2}& 29.8 \textsubscript{14.8}& 28.9 \textsubscript{17.1}& 29.4 \textsubscript{14.6}\\
 & GPT2-XL & 45.6 \textsubscript{9.9}& 40.0 \textsubscript{10.7}& 39.1 \textsubscript{8.4}& 40.6 \textsubscript{9.6}& 35.6 \textsubscript{12.3}& 36.2 \textsubscript{9.2}& 33.0 \textsubscript{14.1}& 33.5 \textsubscript{9.3}& 28.0 \textsubscript{16.7}& 33.5 \textsubscript{11.9}\\
 & GPT-J & 44.8 \textsubscript{10.9}& 47.3 \textsubscript{13.3}& 41.2 \textsubscript{9.4}& 42.0 \textsubscript{12.6}& 36.8 \textsubscript{10.7}& 36.3 \textsubscript{8.5}& 30.5 \textsubscript{12.9}& 30.9 \textsubscript{9.5}& 25.4\textsubscript{14.6} & 29.4 \textsubscript{10.5}\\ \hline
\end{tabular}}
\caption{Performance comparison of supervised learning and in-context learning on noisy labels. We show the mean and standard deviation across different classification types. The subscripts are the standard deviations over 5 different seeds.}
\label{tab:2}
\end{table*}

\begin{table*}[t]
\centering
\begin{tabular}{cccccccc}
\hline
\multirow{3}{*}{\textbf{Dataset}} & \multirow{3}{*}{\textbf{LM}} & \multicolumn{6}{c}{\textbf{Imbalance Ratio}} \\ \cline{3-8} 
 &  & \multicolumn{2}{c}{Low} & \multicolumn{2}{c}{Medium} & \multicolumn{2}{c}{High} \\ \cline{3-8} 
 &  & SL & ICL & SL & ICL & SL & ICL \\ \hline
\multirow{3}{*}{Binary} & GPT2-Large & 52.8 \textsubscript{6.4}& 49.1 \textsubscript{11.3}& 47.8 \textsubscript{6.4}& 49.4 \textsubscript{10.8}& 36.4 \textsubscript{0.3}& 46.1 \textsubscript{11.9}\\
 & GPT2-XL & 51.7 \textsubscript{6.5}& 41.1 \textsubscript{5.9}& 46.4 \textsubscript{4.4}& 52.6 \textsubscript{5.6}& 35.3 \textsubscript{1.5}& 39.3 \textsubscript{4.6}\\
 & GPT-J & 52.4 \textsubscript{6.1}& 52.5 \textsubscript{12.2}& 43.0 \textsubscript{1.7}& 53.6 \textsubscript{17.0}& 34.3 \textsubscript{0.5}& 53.5 \textsubscript{20.2}\\ \hline
\multirow{3}{*}{Multi-class} & GPT2-Large & 48.1 \textsubscript{9.8}& 25.6 \textsubscript{15.4}& 43.8 \textsubscript{15.3}& 23.1  \textsubscript{15.8}& 33.2 \textsubscript{13.9}& 21.8 \textsubscript{11.1}\\
 & GPT2-XL & 44.6 \textsubscript{5.6}& 39.6 \textsubscript{6.8}& 47.0 \textsubscript{12.8}& 35.8 \textsubscript{11.4}& 38.5 \textsubscript{14.8}& 36.4 \textsubscript{11.2}\\
 & GPT-J & 47.4 \textsubscript{11.6}& 48.7 \textsubscript{17.5}& 47.9 \textsubscript{15.4}& 51.8 \textsubscript{17.3}& 36.3 \textsubscript{18.5}& 38.2 \textsubscript{12.2}\\ \hline
 \hline
 \multirow{3}{*}{AVERAGE} & GPT2-Large & 50.5 \textsubscript{7.8}& 37.4 \textsubscript{17.6}& 45.8 \textsubscript{10.7}& 36.3 \textsubscript{18.8}& 34.8 \textsubscript{8.9}& 34.0 \textsubscript{16.8}\\
 & GPT2-XL & 48.2 \textsubscript{6.7}& 40.4 \textsubscript{5.8}& 46.7 \textsubscript{8.5}& 44.2 \textsubscript{12.2}& 36.9 \textsubscript{9.6}& 37.9 \textsubscript{7.8}\\
 & GPT-J & 49.9 \textsubscript{8.7}& 50.6 \textsubscript{13.7}& 45.4 \textsubscript{10.2}& 52.7 \textsubscript{15.3}& 35.3 \textsubscript{11.8}& 45.9 \textsubscript{17.1}\\ \hline
\end{tabular}
\caption{Performance comparison of supervised learning and in-context learning on label imbalance. We show the mean and standard deviation across different classification types. The subscripts are the standard deviations over 5 different seeds.}
\label{tab:3}
\end{table*}

\begin{figure}[t]
\begin{center}
\vspace{-2mm}
\includegraphics[width=\columnwidth]{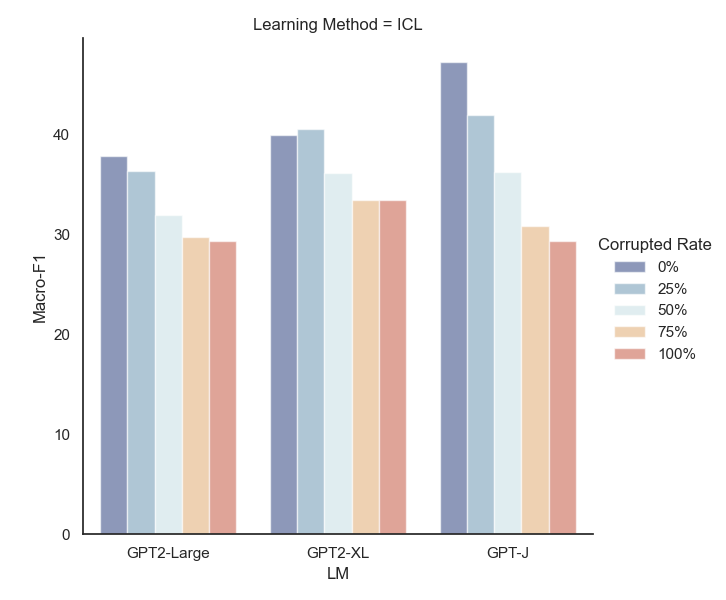}
\vspace{-6mm}
\caption{Results with varying ratio of correct labels for in-context learning over three different models.}\label{fig:2}
\end{center}
\end{figure}

\subsection{How do ICL and SL Learn with Noisy Labels?}
 \paragraph{Ground-truth Labels matter a lot to ICL}  Min {\em et al.}\ \cite{sewon-etal-2022-rethinking}  claimed that replacing ground-truth labels with incorrect labels marginally affects the overall performance on specific datasets. In contrast, our work conducts extensive experiments on the impact of gold labels, and it is noticeable that the ability of ICL emerges differently depending on corruption rates, as shown in Figure \ref{fig:2}. Model performance is sensitive to the ratio of correct labels in the demonstrations. In fact, using all corrupted labels always significantly under-performs demonstration with all gold labels across all three models. Meanwhile, as the model size increases, the performance gap between various corruption rates also increases, which indicates that the large language models are more sensitive to the corrupted labels. For instance, the performance gap between all gold and all corrupted labels for GPT2-Large is 8.5\%, and for GPT-J is 17.9\%. 

Correct labels matter a lot to the model performance but, in  rare cases, model performance is not guaranteed with lower corruption rates. 
An inflection point might exist where a certain number of corrupted labels helps {\em improve} model performance. For instance, 75\% correct labels performs slightly better than using all gold labels for GPT2-XL.

\begin{figure*}[t]
\begin{center}
\includegraphics[width=\textwidth]{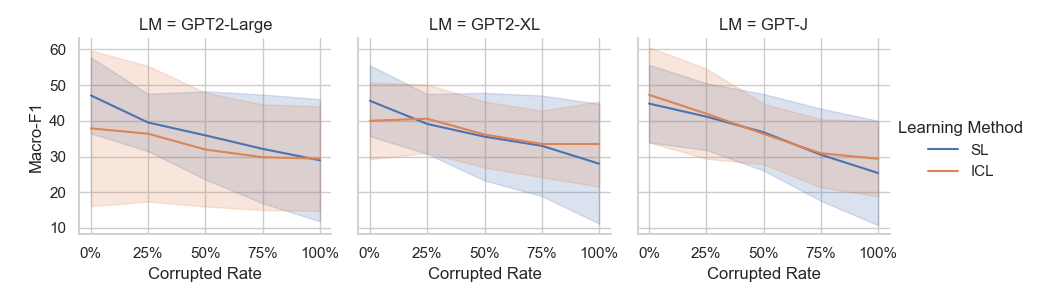}
\vspace{-5mm}
\caption{Performance comparison of supervised learning and in-context learning under different corrupted rates. We plot the mean accuracy ($\pm$ standard deviation) across six different datasets for three different model sizes.}
\label{fig:3}
\end{center}
\end{figure*} 

\begin{table}
\centering
\begin{tabular}{ccccc}
\hline
\multirow{2}{*}{\textbf{Model}} & \multicolumn{2}{c}{\textbf{$\beta_{1}$}}& \multicolumn{2}{c}{\textbf{$\beta_{0}$}} \\ \cline{2-5}
 & \multicolumn{1}{c}{SL} & ICL & \multicolumn{1}{c}{SL} & ICL \\ \hline
GPT2-Large & \multicolumn{1}{c}{-0.174} & -0.094 & \multicolumn{1}{c}{0.454} & 0.378 \\ \hline
GPT2-XL & \multicolumn{1}{c}{-0.165} & -0.080 & \multicolumn{1}{c}{0.445} & 0.407 \\ \hline
GPT-J & \multicolumn{1}{c}{-0.198} & -0.187 & \multicolumn{1}{c}{0.456} & 0.465 \\ \hline
\end{tabular}
\caption{Sensitivity measure on the performance against the different noisy levels}
\label{tab:4}
\end{table}

\paragraph{Compared with SL} Overall, as the corrupted rate increases, the performance of supervised learning declines dramatically, and the performance of ICL drops slowly for GPT2-Large and GPT2-XL but decreases as rapidly as SL for GPT-J, as shown in Figure \ref{fig:3}. Looking at the sensitivity measures in Table \ref{tab:4}, we see that ICL is less sensitive than SL across all three models. In addition, ICL gradually outperforms SL as the model size increases. 
SL performance drops up to 19\% after injecting incorrect labels, while ICL performance drops 11\%.
It is also noticeable that, as more incorrect labels are  introduced, the performance of ICL declines gradually at first and then becomes constant, while SL steadily declines. ICL gradually attains performance comparable to that of SL as the model size increases.

We are also interested in how various learning paradigms perform with different classification types. We divide the six classification tasks into two types, binary classification and multi-class classification. As shown in Table \ref{tab:2}\footnote{The detailed visualization can be found in Figure 1 in the Technical Appendix B at \urlstyle{same}\url{https://github.com/xdwang0726/ICL_LL/blob/main/ECAI_Supplementary_Document.pdf}.}, there is no large gap between the performance of SL and ICL in binary classification. Specifically, in binary classification with GPT2-large and GPT-J, ICL is slightly better with lower corruption rates and SL improves with small advantages when dealing with large corruption rates. SL outperforms ICL slightly across most corruption rates when using GPT2-XL. With multi-class classification, as the number of corrupted label increases, the performance of ICL gradually outperforms SL across all three models. The cause of different tendencies in performance between ICL and SL on binary and multi-class classification may be that binary classification tasks are generally simpler than multi-class tasks. With binary tasks, SL may be able to learn accurate decision boundaries more easily, while ICL may not provide additional benefit beyond the labeled examples. When faced with more complex multi-class tasks, ICL may better leverage the context of the inputs to better understand the relationships between the labels and provide more accurate predictions as the size of the model increases. 

\begin{figure}[t]
\begin{center}
\vspace{-3mm}
\includegraphics[width=\columnwidth]{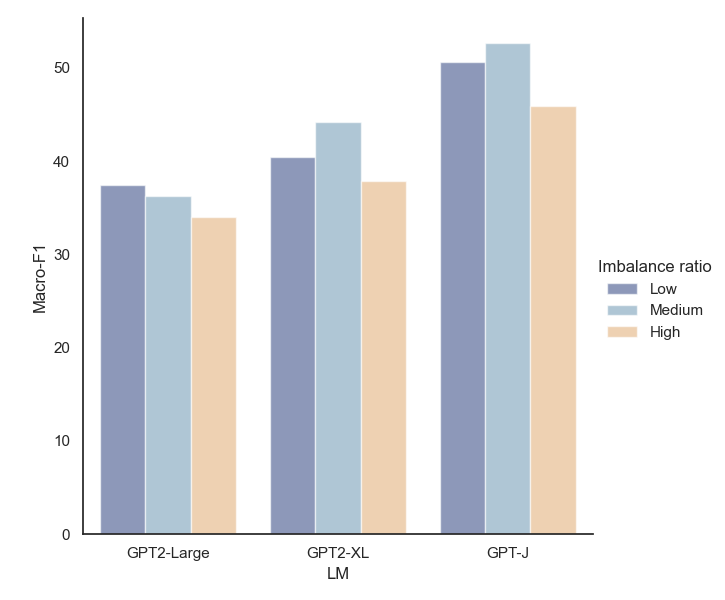}
\vspace{-5mm}
\caption{Results with varying imbalance ratios for in-context learning across three different models}\label{fig:4}
\end{center}
\end{figure} 

\paragraph{Discussion} The learning behaviour of SL is strongly affected by the ground truth input-label correspondence as the model size increases. It suggests that SL requires correct pairings to perform downstream tasks. We hypothesize that, for ICL, the wrong pairings sabotage the pre-trained knowledge to some extent, but the large language models are more able to absorb knowledge directly from the pairings in the demonstrations. Similarly, input-label correspondence has a consistent effect on the learning behaviour of SL across all model sizes, but SL is more sensitive to corrupted labels compared to ICL. The reason for this may be that with SL, the model is fine-tuned by updating parameters using the wrong input-label pairs which enlarge the noisy ``signal'' in the labels. SL may be more prone to over-fitting than ICL, especially with high corruption rates. When the labels are noisy, the model may learn to fit the noise data during training rather than the underlying patterns, which may lead to poor generalization performances. Also, ICL takes into account the context in which the data are presented. In ICL, the model learns to predict the label based on not only the input features but also the context that can help to mitigate the impact of noisy and mislabeled data. 

\subsection{How do ICL and SL Learn with Imbalanced Labels?}
\paragraph{Label Imbalance Matters Little to ICL} As shown in Figure \ref{fig:4}, label imbalance does not have significant effects on ICL. Specifically, decreasing the number of positive examples (i.e., increasing the level of label imbalance) does not lead to a significant drop in performance for any of the models.
We also notice that learning with a low imbalance ratio (a balanced dataset) is not always the best choice, and a certain level of label imbalance may be expected or even desirable. 
For instance, by using GPT2-XL and GPT-J, the performances with medium imbalance ratios slightly outperform those with low imbalance ratios. One possible reason is that, in some cases, the presence of a reasonably high proportion of negative examples may actually help the language model learn to better discriminate between the positive and negative cases. This is because the language model can learn to recognize common patterns or characteristics that are associated with the negative examples, which can then help it more easily identify the positive ones.

\begin{table}
\centering
\begin{tabular}{ccccc}
\hline
\multirow{2}{*}{\textbf{Model}} & \multicolumn{2}{c}{\textbf{$\beta_{1}$}} & \multicolumn{2}{c}{\textbf{$\beta_{0}$}} \\ \cline{2-5}
 & \multicolumn{1}{c}{SL} & ICL & \multicolumn{1}{c}{SL} & ICL \\ \hline
GPT2-Large & \multicolumn{1}{c}{-0.157} & -0.034 & \multicolumn{1}{c}{0.515} & 0.376 \\ \hline
GPT2-XL & \multicolumn{1}{c}{-0.113} & -0.025 & \multicolumn{1}{c}{0.495} & 0.420 \\ \hline
GPT-J & \multicolumn{1}{c}{-0.146} & -0.047 & \multicolumn{1}{c}{0.508} & 0.520 \\ \hline
\end{tabular}
\caption{Sensitivity measure on the performance against the different imbalance ratio}
\label{tab:5}
\end{table}

\begin{figure*}[t]
\begin{center}
\includegraphics[width=\textwidth]{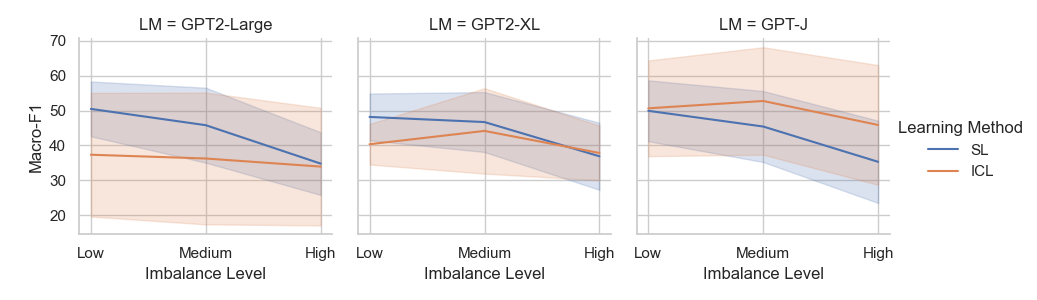}
\vspace{-5mm}
\caption{Performance comparison of supervised learning and in-context learning under different imbalance ratios. We plot the mean accuracy ($\pm$ standard deviation) across six different datasets for three different model sizes.}
\label{fig:5}
\end{center}
\end{figure*}

\paragraph{Compared with SL} The sensitivity measure in Table \ref{tab:5} shows that ICL is less sensitive to label imbalance compared to SL consistently across all three models. This indicates that SL is vulnerable to imbalanced labels, and the performance drops up to 15\% from relatively balanced labels to highly imbalanced ones on average across three models. However, ICL is not so sensitive, with performance differences fluctuating between 3.4\% and 6.8\% over imbalance ratios. In addition, Figure \ref{fig:5} shows that the performance of ICL gradually surpasses SL as the model size increases. 

In Table \ref{tab:3}\footnote{Detailed visualization can be found in Figure 2 in the Technical Appendix B at \urlstyle{same}\url{https://github.com/xdwang0726/ICL_LL/blob/main/ECAI_Supplementary_Document.pdf}.}, we also observe that, with binary classification, ICL shows sustained sensitivity as the imbalance ratio increases. ICL always outperforms supervised learning with medium and high imbalance ratios across all three models. With multi-class classification, though ICL is less sensitive to label imbalance, the model performance is not superior to that of SL, especially with GPT2-Large and GPT2-XL. With GPT-J, the advantage of ICL becomes obvious, which indicates that, with large language models, ICL is clearly preferable when dealing with class-imbalanced datasets.  

\paragraph{Discussion} In SL, imbalanced data imposes ``label bias'' during learning where the decision boundaries can be strongly influenced by the frequent classes. SL is incented to learn features from the frequent classes and may ignore other features. In contrast, ICL achieves non-trivial performance for classification tasks with imbalanced demonstrations as it may mitigate the effects of bias by leveraging the context specific information. This strongly suggests that the learning behaviour of ICL is not affected by the number of instances per class in the demonstrations, and ICL is capable of preserving pre-training knowledge as well as learning richer features. The reason for this may be that ICL can learn not only the label-relevant features but also other interesting features that capture the intrinsic properties of the input distributions and the structures of the input-label correspondence, which may generalize better to rare classes. This hypothesis needs further study.

\section{Analysing ICL's Sensitivity Using Attention Score} 
Our findings seem intuitive but are not trivial. The language models are not pre-trained on perturbed samples, but still ICL is less sensitive to label perturbations compared to SL. Understanding these observations are crucial for practitioners to apply ICL. The attention mechanism \cite{bahdanau2014neural} has been used as an important component across LLMs. An attention layer produces a distribution over input representations to be attended to, and the distribution can reflect the significance of the inputs. Inspired by Wiegreffe and Pinter \cite{wiegreffe-pinter-2019-attention} that shows the utility of attention scores in interpretation, we further examine why ICL is less sensitive to label perturbations using attention scores. Specifically, we compute the attention scores of the output label word in each example for demonstration $\mathcal{C}$ and sum up the scores for the same class together. We report the summed attention score ratio across different class labels (i.e. correct labels and perturbed labels) in the MR dataset and refer this as \emph{Attention ratio score} that explains the robustness of ICL under the imbalance and noisy label settings. 

\begin{table}

\resizebox{\columnwidth}{!}{
\begin{tabular}{cccccccc}
\hline
\multirow{2}{*}{\textbf{LM}} & \multicolumn{6}{c}{\textbf{Ratios (corrected : corrupted)}} \\ \cline{2-8} 
 & \multicolumn{1}{c}{5:0} & \multicolumn{1}{c}{4:1} & \multicolumn{1}{c}{3:2} & \multicolumn{1}{c}{2:3} & \multicolumn{1}{c}{1:4} & \multicolumn{1}{c|}{0:5} & STD \\ \hline
GPT2-Large & \multicolumn{1}{c}{0.424} & \multicolumn{1}{c}{0.428} & \multicolumn{1}{c}{0.434} & \multicolumn{1}{c}{0.443} & \multicolumn{1}{c}{0.459} & \multicolumn{1}{c|}{0.453} & 0.014\\ \hline \hline
GPT2-XL & \multicolumn{1}{c}{0.521} & \multicolumn{1}{c}{0.561} & \multicolumn{1}{c}{0.525} & \multicolumn{1}{c}{0.591} & \multicolumn{1}{c}{0.592} & \multicolumn{1}{c|}{0.564} & 0.031\\ \hline \hline
GPT-J & \multicolumn{1}{c}{0.370} & \multicolumn{1}{c}{0.393} & \multicolumn{1}{c}{0.428} & \multicolumn{1}{c}{0.297} & \multicolumn{1}{c}{0.323} & \multicolumn{1}{c|}{0.316} & 0.051\\ \hline 
\end{tabular}}
\caption{Attention ratio scores for different correct label ratios for three models. Scores between models are not comparable, as attention scores vary. \emph{STD} is the standard deviation across different corrupted ratios per model.} 
\label{tab:6}
\end{table}


\paragraph{Noisy Labels} To construct the experiment with noisy labels, we use 5 demonstration examples ($k=5$) and evaluate with various corrupted rates. 
Results are shown in Table \ref{tab:6}. With relatively small language models, such as GPT2-Large and GPT2-XL, there are small fluctuations of attention scores if more corrupted labels are added to the demonstration; however, with larger models like GPT-J, a large variation of attention  appears as the number of corrupted labels increases. The aforementioned observation further indicates that smaller models cannot differentiate gold labels and corrupted ones and pay approximately the same attention to all labels; however, large language models demonstrate an ability to disambiguate between gold and corrupted labels, which supports our suspicion.  
\paragraph{Label Imbalance} To construct the experiment with class-imbalanced data, we evaluate the attention score ratio of the two classes by fixing the number of examples of one class and adding examples for another. We report their attention score ratio in Table \ref{tab:7}. The attention scores change slightly with different imbalance ratios, which indicates that each class receives approximately the same attention from ICL even with imbalanced label distributions.

\begin{table}
\resizebox{\columnwidth}{!}{
\begin{tabular}{cccccccc}
\hline
\multirow{2}{*}{\textbf{LM}} & \multicolumn{6}{c}{\textbf{Imbalance Ratio}} \\ \cline{2-8}
 & \multicolumn{1}{c}{100\%} & \multicolumn{1}{c}{50\%} & \multicolumn{1}{c}{33\%} &  \multicolumn{1}{c}{25\%} &  \multicolumn{1}{c}{20\%} & \multicolumn{1}{c|}{10\%} & STD\\ \hline
GPT2-Large & \multicolumn{1}{c}{0.946} & \multicolumn{1}{c}{0.956} & \multicolumn{1}{c}{0.936} & \multicolumn{1}{c}{0.967} & \multicolumn{1}{c}{0.913} & \multicolumn{1}{c|}{0.887} & 0.030\\ \hline \hline
GPT2-XL & \multicolumn{1}{c}{0.929} & \multicolumn{1}{c}{0.995} & \multicolumn{1}{c}{0.983} & \multicolumn{1}{c}{0.995} & \multicolumn{1}{c}{0.913} & \multicolumn{1}{c|}{0.975} & 0.035\\ \hline \hline
GPT-J & \multicolumn{1}{c}{0.909} & \multicolumn{1}{c}{0.825} & \multicolumn{1}{c}{0.847} & \multicolumn{1}{c}{0.857} & \multicolumn{1}{c}{0.908} & \multicolumn{1}{c|}{0.913} & 0.038\\ \hline
\end{tabular}}
\caption{Attention ratio scores for different imbalance ratios for three models. Scores between models are not comparable, as attention scores vary. \emph{STD} is the standard deviation across different imbalance ratios per model.}
\label{tab:7}
\end{table}

\section{Related Work}

\subsection{Learning with Large Pre-trained Language Models}

Large pre-trained language models have become a cornerstone of natural language processing, and have shown strong generalization abilities on a wide range of tasks \cite{NEURIPS2020_1457c0d6, devlin-etal-2019-bert, radford2019language, JMLR:v21:20-074}. Currently, there are two key frameworks \cite{https://doi.org/10.48550/arxiv.2205.05638} to use LLMs, namely supervised fine-tuning and in-context learning. 

Supervised learning performs gradient-based fine-tuning to adapt a task-agnostic pre-trained model for a specific task. Existing work \cite{https://doi.org/10.48550/arxiv.2002.06305, https://doi.org/10.48550/arxiv.1909.11299, Phang2018SentenceEO} shows that supervised fine-tuning significantly boosts the performance on various NLP tasks, especially on small datasets. 

In-context learning learns a new task without updating the model's parameters. ICL adds a prompt (usually a task description) and annotated examples as demonstrations (also known as ``shots'') to enable few-shot learning without fine-tuning \cite{NEURIPS2020_1457c0d6}. Some recent work has been devoted to investigating ICL. For instance, Min {\em et al.}\ \cite{sewon-etal-2022-rethinking} tried to explain and understand the mechanism of ICL and found correct input-label mapping matters very little, while later work \cite{https://doi.org/10.48550/arxiv.2205.12685} further argued the impact of correct mappings under different configurations. Zhao {\em et al.}\ \cite{pmlr-v139-zhao21c} showed that models are biased by the order of the demonstration examples. Liu {\em et al.}\ \cite{liu-etal-2022-makes} and Mishra {\em et al.}\ \cite{mishra-etal-2022-reframing} focused on the sensitivity of ICL: the former empirically studied the sensitivity of GPT-3's few-shot capabilities with respect to the selection of in-context examples and the latter re-framed the instructional prompts to improve the in-context performance for language models. Chen {\em et al.}\ \cite{chen-etal-2022-meta} and Min {\em et al.}\ \cite{min-etal-2022-metaicl} used meta-training with an explicit in-context learning objective. Xie {\em et al.}\ \cite{https://doi.org/10.48550/arxiv.2111.02080} and Aky\"{u}rek {\em et al.}\ \cite{akyurek2022learning} studied what enables ICL: the former explained ICL as implicit Bayesian inference and the latter used linear regression as a prototypical problem to study the learning algorithm of ICL. Yoo {\em et al.}\ \cite{https://doi.org/10.48550/arxiv.2205.12685} shared the same conclusion that ground-truth labels matter to ICL, as we claim, but by conducting experiments only on GPT-J. Our experiments cover three models with various sizes and we draw the more in-depth conclusion that LLMs are more sensitive to corrupted labels than smaller models. Parallel to our research, Zhang {\em et al.}\ \cite{zhang-etal-2022-active} claimed that a well-balanced demonstration set does not consistently lead to better performance in ICL by studying the class imbalance under the demonstration example selection. While we reach a similar conclusion, we saw the result that label imbalance matters little to ICL when comparing ICL and SL. 

\subsection{Training with Perturbed Labels} 
\paragraph{Noisy Labels} 
Learning with noisy labels can be divided into attempts that handle noisy labels in data pre-processing and attempts to learn robustly from the noisy labels directly. Some approaches pre-process auxiliary data by selecting clean instances according to rules. For instance, Li {\em et al.}\ \cite{8237473} learned a teacher network with a small clean dataset to do importance re-weighting of the noisy labels in the loss function. Similarly, MentorNet \cite{pmlr-v80-jiang18c} learned a data-driven curriculum to select clean instances that could guide the training of a student network. Further, Chen {\em et al.}\ \cite{pmlr-v97-chen19g} applied cross-validation to randomly split the noisy dataset and remove large-loss samples. 
Other approaches alleviate noise by making the learning process more robust to label noise, mainly among three categories: robust architectures \cite{7837934, pmlr-v97-lee19f, 7298885}  add a noise adaptation layer to learn label transitions or develop a dedicated architecture to support diverse types of label noise; robust regularization \cite{Menon2020CanGC, Pereyra2017RegularizingNN} that enforces a deep neural network to overfit less to false-labeled examples explicitly or implicitly; and robust loss functions \cite{10.5555/3327546.3327707, 7929355} which adjust the loss  according
to the confidence of the labels. 

\paragraph{Label Imbalance}  Imbalanced text classification focuses mainly on two types of methods:  sampling-based and algorithm-level. Sampling-based methods balance  class distributions by manipulating training examples, including under-sampling, over-sampling , and a mix. Synthetic Minority Oversampling Technique (SMOTE) \cite{SMOTE} is an over-sampling method where synthetic samples are generated for the minority class using the feature space to generate new instances with  linear interpolation between  positive instances. NearMiss \cite{Zhang03} is an under-sampling technique that balances the class distribution by randomly eliminating majority class instances. However, over-sampling can potentially lead to over-fitting, while under-sampling may cause information loss. The algorithm-level approaches pursue balancing the class distribution and lifting the importance of minority class samples without altering the training data distribution, including cost-sensitive learning and threshold-adjustment methods. Cost-sensitive learning modifies cost functions \cite{8237586, 7727770} and thresholding methods change the decision threshold at test time after training the classifier on the original imbalanced data \cite{Ling2008CostSensitiveLA}. 

\section{Conclusion and Future Work}
Our work is the first to study ICL among label perturbations. We empirically show that the learning behaviour of ICL is more influenced by the quality of the ground truth input-label pairings, and it is less effected by the number of pairings for each class in the demonstrations. In comparison with supervised learning (SL), we discover that ICL is less sensitive to perturbed labels than SL, which is still a standard method, and is still useful if the language model is relatively small. However, with large language models, the utility of ICL emerges. These findings highlight the importance of considering different learning strategies (ICL or SL) and model sizes when working with large language models and different classification tasks. Our work encourages practitioners to use ICL instead of SL or, at least, to consider evaluating the impact of perturbed pre-training on downstream tasks if large language models are available (here, we expected better performance from ICL and at least the model size should be larger than GPT-J). Our experiments are limited to text classification tasks, and we have not examined the performance of ICL for other tasks such as text generation nor in other domains such as computer vision. Extending our current work to such tasks is non-trivial, and is left for future work. To understand the learning behaviours of in-context learning and supervised learning, we train and fine-tune a model multiple times with different seeds on a fixed number of demonstration examples, which can be computationally intensive if the sizes of the datasets and models are large (for instance, we fine-tune the GPT-J model on the MR dataset using 4 NVIDIA A100 80G GPUs with parallel computing). This can be a non-trivial problem for practitioners with limited computational resources. Our experiments only have been conducted on the models that range from 774 million to 6.7 billion parameters due to these considerations. As yet, there does not seem to be any evidence that larger LLMs will reverse the tendencies we observed. We hope our findings of the behaviours of ICL under label perturbations can be a useful reference for potential users who have limited computational resources. We would also be interested to explore ways to optimize the learning behaviour of LLMs in different scenarios. We hope our work can inspire analysis of ICL in broader environments in the wild and provide insights for the design of future uses of ICL.

\ack 
We would like to thank all reviewers for their comments, which helped improve this paper considerably. This research is partially funded by The Natural Sciences and Engineering Research Council of Canada (NSERC) through a Discovery Grant to R. E. Mercer. F. Rudzicz is supported by a CIFAR Chair in AI.

\bibliography{ecai}

\end{document}